%% file: 0-main.tex
\documentclass[conference]{IEEEtran}
\IEEEoverridecommandlockouts
\usepackage{cite}
\usepackage{url}
\usepackage{threeparttable}
\usepackage{amsmath,amssymb,amsfonts,multirow,algorithm,algorithmic,float}
\usepackage{algorithmic}
\usepackage{graphicx}
\usepackage{subfigure}
\usepackage{textcomp}
\usepackage{xcolor}

\def\BibTeX{{\rm B\kern-.05em{\sc i\kern-.025em b}\kern-.08em
    T\kern-.1667em\lower.7ex\hbox{E}\kern-.125emX}}
    
\begin{document}

\title{Integrative Analysis of Patient Health Records and Neuroimages via Memory-based Graph Convolutional Network\\
}

\author{
\IEEEauthorblockN{Xi Sheryl Zhang, Jingyuan Chou, Fei Wang}
\IEEEauthorblockA{Weill Cornell Medical College, Cornell University\\
sheryl.zhangxi@gmail.com, jic2015@med.cornell.edu, few2001@med.cornell.edu}
}

\maketitle

\begin{abstract}
With the arrival of the big data era, more and more data are becoming readily available in various real world applications and those data are usually highly heterogeneous. Taking computational medicine as an example, we have both Electronic Health Records (EHR) and medical images for each patient. For complicated diseases such as Parkinson's and Alzheimer's, both EHR and neuroimaging information are very important for disease understanding because they contain complementary aspects of the disease. However, EHR and neuroimage are completely different. So far the existing research has been mainly focusing on one of them. In this paper, we proposed a framework, Memory-Based Graph Convolution Network (MemGCN), to perform integrative analysis with such multi-modal data. Specifically, GCN is used to extract useful information from the patients' neuroimages. The information contained in the patient EHRs before the acquisition of each brain image is captured by a memory network because of its sequential nature. The information contained in each brain image is combined with the information read out from the memory network to infer the disease state at the image acquisition timestamp. To further enhance the analytical power of MemGCN, we also designed a multi-hop strategy that allows multiple reading and updating on the memory can be performed at each iteration. We conduct experiments using the patient data from the Parkinson's Progression Markers Initiative (PPMI) with the task of classification of Parkinson's Disease (PD) cases versus controls. We demonstrate that superior classification performance can be achieved with our proposed framework, comparing with existing approaches involving a single type of data. 

\end{abstract}


\input{1-introduction}

\input{2-method}

\input{3-experiment}

\input{4-relatedwork}
\input{5-conclusion}

\input{6-acknowledgment}

\bibliographystyle{./IEEEtran}
\bibliography{./myrefs}\tiny

\end{document}

%% file: 1-introduction.tex
\section{Introduction}
With the arrival of the big data era, more and more data are becoming readily available in various real world applications. Those data are like gold mines and data mining technologies are like tools that can dig the gold out from those mines. Taking medicine as an example, we have a large amount of medical data of different types nowadays, from molecular to cellular to clinical and even environmental. As has been envisioned in \cite{collins2015new}, one key aspect of precision medicine, which aims at recommending the right treatment to the right patient at the right time, is to integrate those multi-scale data from different sources to obtain a comprehensive understanding of a health condition. 

Many data mining approaches have been proposed for analyzing medical data in recent years. For example, Ghassemi {\em et al.} \cite{ghassemi2014unfolding} modeled the mortality risk in intensive care unit with latent variable models. Caruana {\em et al.} \cite{caruana2015intelligible} utilized generalized additive model to predict the risk of pneumonia and hospital readmission. Zhou {\em et al.} \cite{zhou2014micro} developed a matrix factorization approach for predictive modeling of the disease onset risk based on patients' Electronic Health Records (EHR) data. Tensor modeling techniques have also been leveraged in electronic phenotyping \cite{ho2014marble,wang2015rubik} and clinical natural language processing \cite{luo2015subgraph}. More recently, deep learning has emerged as a powerful data mining approach that can disentangle the complex interactions among data features and achieve superior performance. Because of the complex nature of medical problems, researchers have also been exploring the applicability of deep learning models in helping with medical problems using medical images \cite{esteva2017dermatologist,gulshan2016development},  EHRs \cite{rajkomar2018scalable,zhou2014micro}, physiological signals \cite{rajpurkar2017cardiologist,schirrmeister2017deep}, etc., and obtained promising results.

Despite the initial success, so far most of the existing works on data mining for medicine have been focusing on one single type of data (e.g., images or EHRs). However, typically different data sources contain complementary information about the patients from different aspects. For example, concerning neurological diseases, we can get general clinical information of patients, such as diagnosis, medication, lab, etc., from EHRs; while we can obtain specific biomarkers regarding white matter, gray matter, and the change of different Regions-of-Interest (ROI), from brain images. Integrative analysis of both EHR and neuroimages can help us understand the disease in a better and more comprehensive way. In reality, such integrative analysis is challenging because of the following reasons.
\begin{itemize}
\item {\em Heterogeneity}. The nature of patient EHR and neuroimages are completely different: the EHR for each patient can be regarded as a temporal event sequence, where at each timestamp multiple medical events (e.g., diagnosis, medications, lab tests, etc.) can appear; while each neuroimage is essentially a collection of pixels. Therefore the ways to process these two types of data could be very different.

\item {\em Sequentiality}. EHR data are sequential and a specific brain image is static. The brain status reflected in a certain brain image can be related to the EHR of the corresponding patient before the acquisition of the image. Effective integration of such heterogeneous information into a unified analytics pipeline is a challenging task.
\end{itemize}
With the above considerations, we proposed a novel Memory-based Graph Convolutional Network (MemGCN) to perform integrative analysis with both patient EHRs and neuroimages. As its name suggests, there are two major components in MemGCN.
\begin{itemize}
\item Graph Convolutional Network (GCN) \cite{kipf2016semi}. GCN is a deep learning model that generalizes the Convolutional Neural Nets (CNN) \cite{lecun2015deep} on regular lattices to irregular graphs. GCN has been proved to be very effective on extracting useful features from graphs.

\item Memory Network \cite{weston2015memory}. Memory network is a new type of model that connects a regular learning process with a memory module, which is usually represented as a matrix that memorizes the historical status of the system. At each iteration some useful information is extracted from the memory to help the current inference while the same time the memory unit will be updated.
\end{itemize}
In our framework, the GCN module extracts features from the human brain networks constructed from the brain images. The longitudinal patient EHRs are stored in the memory network to encode the historical clinical information about the patient before the acquisition of the image. The information extracted from the memory network will be combined with the feature from GCN to discriminate PD cases and controls. We conduct experiments on real world data from the patients in the Parkinson's Progression Markers Initiative (PPMI) \cite{marek2011parkinson} and obtained superior performance comparing with conventional methodologies.

The rest of this paper is organized as follows. Section II presents the technical details of our framework. The experimental results are introduced in Section III, followed by the related work in Section IV and conclusions in Section V.

%% file: 2-method.tex
\section{Method}
\begin{figure*}[htb]
    \centering
    \includegraphics[width=7in]{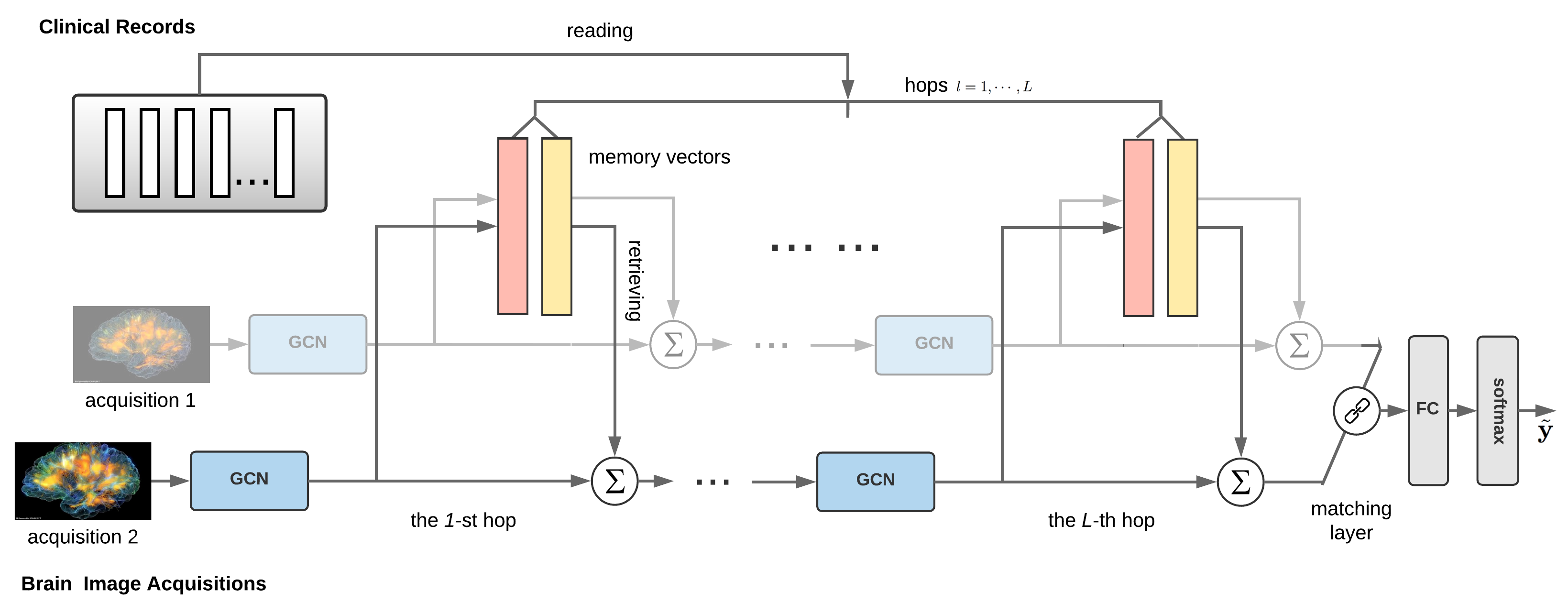}
    \caption{Memory-based Graph Convolutional Network for brain connectivity graphs with clinical records. For simplicity, we depict the clinical records via a sequence of vectors in the figure. In practice, each clinical sequence is corresponding to a neuroimage acquisition.}
    \label{fig:overall}
\vspace{-0.3cm}
\end{figure*}

\subsection{Overview}
As illustrated in Fig.~\ref{fig:overall}, the proposed method MemGCN is a matching network that is designed for metric learning on not only brain images but also clinical records. The preprocessed brain connectivity graphs are transformed by graph convolutional networks into representations, while memory mechanism is in charge of iteratively (multiple hops) reading clinical sequences and choosing what to retrieve from memories in order to augment the representations learned by graph convolution. For the purpose of metric learning, inner product similarity and bilinear similarity are separately introduced in the matching layer. The output component is composed of a fully connected layer and a softmax for relationship classification of acquisition pairs. Accordingly, we present MemGCN, a matching network embeds multi-hop memory-augmented graph convolutions and can be trained in an end-to-end fashion with stochastic optimization. 

\subsection{Graph Convolution}

The brain connectivity graph is characterized by defining its ROI nodes and the interactions among them. Since the graph-structured data are non-Euclidean, it is not straightforward to use a standard convolution that has impressive performances on grid.  
Hence, we resort to geometric deep learning approaches~\cite{bronstein2017geometric, monti2017geometric} to deal with the problem of learning features on brain connectivity network. 

In general, let $\mathcal{G}=(\{1,\cdots,n\}, \mathcal{E}, \mathbf{W})$ be an undirected weighted graph, where $\mathbf{W}=(w_{ij})$ is a symmetric adjacency matrix satisfying $w_{ij}>0$ if $(i,j)\in\mathcal{E}$ and $w_{ij}=0$ if $(i,j)\notin\mathcal{E}$. According to spectral graph theory~\cite{chung1997spectral}, the graph Laplacian matrix can be computed as $\mathbf{\mathbf{\Delta}}=\mathbf{I}-\mathbf{D}^{-1/2}\mathbf{W}\mathbf{D}^{-1/2}$, where $\mathbf{D} \in \mathbb{R}^{n \times n}$ is the diagonal degree matrix with $d_{ii} = \sum_{j\neq i} w_{ij} $, and $\mathbf{I} \in \mathbb{R}^{n \times n}$ is the identity matrix. Note that $\mathbf{\Delta}$ is a positive-semidefinite matrix and its eigendecomposition can be written as $\mathbf{\mathbf{\Delta}} = \mathbf{\Phi}\boldsymbol{\Lambda}\mathbf{\Phi}^{\mathrm{T}}$, where 
$\mathbf{\Phi}=(\phi_{0},\cdots,\phi_{n})$ are the orthonormal eigenvectors and $\boldsymbol{\Lambda}=diag(\lambda_{1},\cdots,\lambda(n))$ is the diagonal matrix of non-negative eigenvalues $0=\lambda_{0}\leq\cdots\leq\lambda_{n}$.   

In our scenario, the vertices of graph $\mathcal{G}$ are corresponding to ROIs. Define a brain connectivity acquisition as an input signal $\mathbf{x}=(\mathbf{x}_{1},\cdots,\mathbf{x}_{n})$, where $\mathbf{x}_{i}\in\mathbb{R}^{n}$ is a feature vector associated with vertex $i$. The convolution operation is conducted on Fourier domain instead of the vertex domain. Consider two signals $\mathbf{x}$ and $\mathbf{g}$, it can be proved that the following equation exists, 
\begin{align}
\mathbf{x}\star\mathbf{g}=\mathbf{\Phi}(\mathbf{\Phi}^{\mathrm{T}}\mathbf{x})\odot(\mathbf{\Phi}^{\mathrm{T}}\mathbf{g})=\mathbf{\Phi}g_{\theta}(\boldsymbol{\Lambda})\mathbf{\Phi}^{\mathrm{T}}\mathbf{x} \nonumber \\ 
=\mathbf{\Phi}diag(\hat{g}_{1},\cdots\hat{g}_{n})\hat{\mathbf{x}} \end{align}
where $\odot$ is the element-wise Hadamard product and $\hat{\mathbf{x}}=\mathbf{\Phi}^{\mathrm{T}}\mathbf{x}$ defines the Graph Fourier Transform. The function $\hat{g}_{\theta}(.)$ can be regarded as learnable spectral filters. Previous studies~\cite{bruna2014spectral,henaff2015deep,defferrard2016convolutional,kipf2016semi} on geometric deep learning have proposed a variety of filter functions to achieve promising properties such as spatial localization and computational complexity. Chebyshev spectral convolution network (ChebNet)~\cite{defferrard2016convolutional} is utilized in our model. Before introducing representation learning by ChebNet, we first give the details about how to construct a graph $\mathcal{G}$ and build its edges $\mathcal{E}$ with ROI vertices of the collection of brain image acquisitions.      

\noindent{\bf Spatial Graph Construction}

\label{sec:gcn}
The brain connectivity graph can be represented as a square matrix $\mathbf{x}\in\mathbb{R}^{n\times n}$ with the numerical values indicating the connectivity strength of ROI pairs. However, the region coordinates of anatomical space can provide the crucial spatial relations between ROIs which have not been taken into account in conventional works of the domain~\cite{kawahara2017brainnetcnn}. Motivated by the work~\cite{ktena2017distance}, which applied graph convolution on a functional Magnetic Resonance Imaging (fMRI) task, a spatial graph based on 3-dimensional coordinates is constructed for our model. The coordinates are associated with a predefined number of ROIs and share a common coordinate system. 

In detail, the xyz-coordinates $\{(v^{x}_{i,m}, v^{y}_{i,m}, v^{z}_{i,m})\}_{m=1}^M$ of region center are able to present the spatial location of the corresponding ROI $i$. The global ROI coordinates are computed by the average aggregation $\bar{v}_{i}=\frac{1}{M}(\Sigma_{m}^{M}v^{x}_{i,m}, \Sigma_{m}^{M}v^{y}_{i,m}, \Sigma_{m}^{M}v^{z}_{i,m}), \forall i \in (1,\cdots,n)$. Thus, the edges $\mathcal{E}$ can be constructed by a Gaussian function based on $k$-Nearest Neighbor similarity, which is
\begin{equation}
w_{ij} =
        \begin{cases}
            \text{exp} (-\frac{\|\bar{v}_{i} - \bar{v}_{j} \|^2}{2 \sigma^2}), & \text{if}~~i \in \mathcal{N}_j ~\text{or}~j \in \mathcal{N}_i  \\
            0, & \text{otherwise}. 
        \end{cases}
\end{equation}
where $w_{ij}$ denotes the edge weights between vertex $i$ and vertex $j$, $\mathcal{N}_i$ and $\mathcal{N}_j$ denote the neighbors for $i$ and $j$ respectively. In practice, we set $\mathcal{G}$ as a $10$-Nearest Neighbor graph. Therefore, the spatial information of ROI is formulated into our model in terms of the graph structure.        

\noindent{\bf ChebNet}

With the constructed graph $\mathcal{G}$, its graph Laplacian matrix $\mathbf{\Delta}$ can be obtained. Now our goal is to learn a high-level representation for each image acquisition by feeding its input signal $\mathbf{x}$ as well as the shared $\mathbf{\Delta}$ into the neural network. From the general sense, it can capture the local traits of each individual brain images and the global traits of the population of subjects. 

To address the issues of localization and computational efficiency for  convolution filters on graphs, ChebNet exploited a series of polynomial filters represented in the Chebyshev basis,
\begin{equation} \label{eq:GC}
	g_{\theta}(\mathbf{\Delta})=\sum_{p=0}^{r-1}\theta_{p}T_{p}(\tilde{\mathbf{\Delta}})=\sum_{p=0}^{r-1}\theta_{p}\mathbf{\Phi}T_{p}(\tilde{\mathbf{\Lambda}})\mathbf{\Phi}^{\mathrm{T}}
\end{equation}
where $\tilde{\mathbf{\Delta}} = 2\lambda_{n}^{-1}\mathbf{\Delta} - \mathbf{I}$ is the rescaled Laplacian which leads to its eigenvalues $\tilde{\mathbf{\Lambda}} = 2\lambda_{n}^{-1} \mathbf{\Lambda} - \mathbf{I}$ in the interval $[-1, 1]$. $\mathbf{\theta}$ is the $r$-dimensional vector Chebyshev coefficients parameterizing the filters.  
And $T_p(\tilde{\lambda})=2\lambda T_{j-1}(\lambda)-T_{j-2}(\lambda)$ defines the Chebyshev polynomial in a recursive manner with $T_{0}(\lambda)=1$ and $T_{1}(\lambda)=\lambda$.

To explicitly express filter learning of the graph convolution, without loss of generality, let $k^{l}$ denote the index of feature map in layer $l$, the $k^{l+1}$-th feature map in its layer of sample $m$ is given by     
\begin{equation}
\mathbf{y}_{m,k^{l+1}} = \sum_{k^{l}=1}^{f_{in}} g_{\theta_{k^{l,l+1}}} (\mathbf{\Delta}) \mathbf{y}_{m,k^{l}} \in \mathbb{R}^n 
\end{equation}
yielding $f_{in} \times f_{out}$ vectors of trainable Chebyshev coefficients $\theta_{k^{l,l+1}} \in \mathbb{R}^r$. In detail, $\mathbf{y}_{m,k^{l}}$ denotes the feature maps of the $l$-th layer. For the input layer, $\mathbf{y}_{m,k^{l}}$ can be simply set as $\mathbf{x}_{m,i}, ~i=1,\cdots,n$, which is the $i$-th row vector of the $n\times n$ brain connectivity matrix. Given a sample $\mathbf{x}\in\mathbb{R}^{n \times n}$, its output of graph convolution can be collected into a feature matrix $\mathbf{y} = (\mathbf{y}_1, \mathbf{y}_2, \cdots, \mathbf{y}_{f_{out}}) \in \mathbb{R}^{n \times f_{out}}$, where each row represents the learned high-level feature vector of an ROI.

\subsection{Memory Augmentation}
\label{sec:memory}
The key contribution of MemGCN is incorporating sequential records into the representation learning of brain connectivity in terms of memories. Our model is proposed based on Memory Networks~\cite{weston2015memory,sukhbaatar2015end} which has a variety of successful uses in natural language processing tasks~\cite{bordes2015large,miller2016key,graves2014neural} including complex reasoning or question answering. When we define a memory, it could be viewed as an array of slots that can encode both long-term and short-term context. By pushing the clinical sequences into the memories, the continuous representations of this external information are processed with brain graphs together so that a more comprehensive diagnosis could be made. Inspired by the observation, the memory-augmented graph convolution are designed. 

We start by introducing the MemGCN in the single hop operation, and then show the architecture of stacked hops in multiple steps. Concretely, the memory augmentation can be divided into two procedures: reading and retrieving(see Fig.~\ref{fig:overall}).

\noindent{\bf Clinical Sequences Reading}

Suppose there is a discrete input clinical sequences $\mathbf{s}_{j}, j=1,\cdots,t$, where $j$ is the index of a clinical record extracted from the certain timestamp. In memory network, it needs to be transformed as continuous vectors $\mathbf{z}_{j}, ~j=1,\cdots,t$ and stored into the memory. We use a fixed number of timestamps $t$ to define the memory size. The dimension of the continuous space is denoted as $d$ while the dimension of the original clinical features is denoted as $D$. To embed the sequential vectors $\mathbf{s}_{1},\cdots\mathbf{s}_{t}$, a $d\times D$ embedding matrix $\mathbf{A}$ is used. That is $\mathbf{z}_{j}=\mathbf{A}\mathbf{s}_{j}$. The matrix $\mathbf{z}=(\mathbf{z}_{1},\cdots,\mathbf{z}_{t})$ can be regarded as a new input memory representation. 

Meanwhile, similar to the method in~\cite{sukhbaatar2015end}, an output memory to generate continuous vectors $\{\mathbf{e}_{j}\}$ is involved. The corresponding embeddings are obtained from $\mathbf{e}_{j}=\mathbf{B}\mathbf{s}_{j}$, where $\mathbf{B}$ also is a $d\times D$ embedding matrix. Different from other computational forms of attentive weights\cite{bahdanau2014neural}, two memories in our model are maintained by the separate sequence reading procedures, which are responsible for memory access and integration respectively in the retrieving procedure.  

\begin{figure}[t]
    \centering
    \includegraphics[width=3.5in]{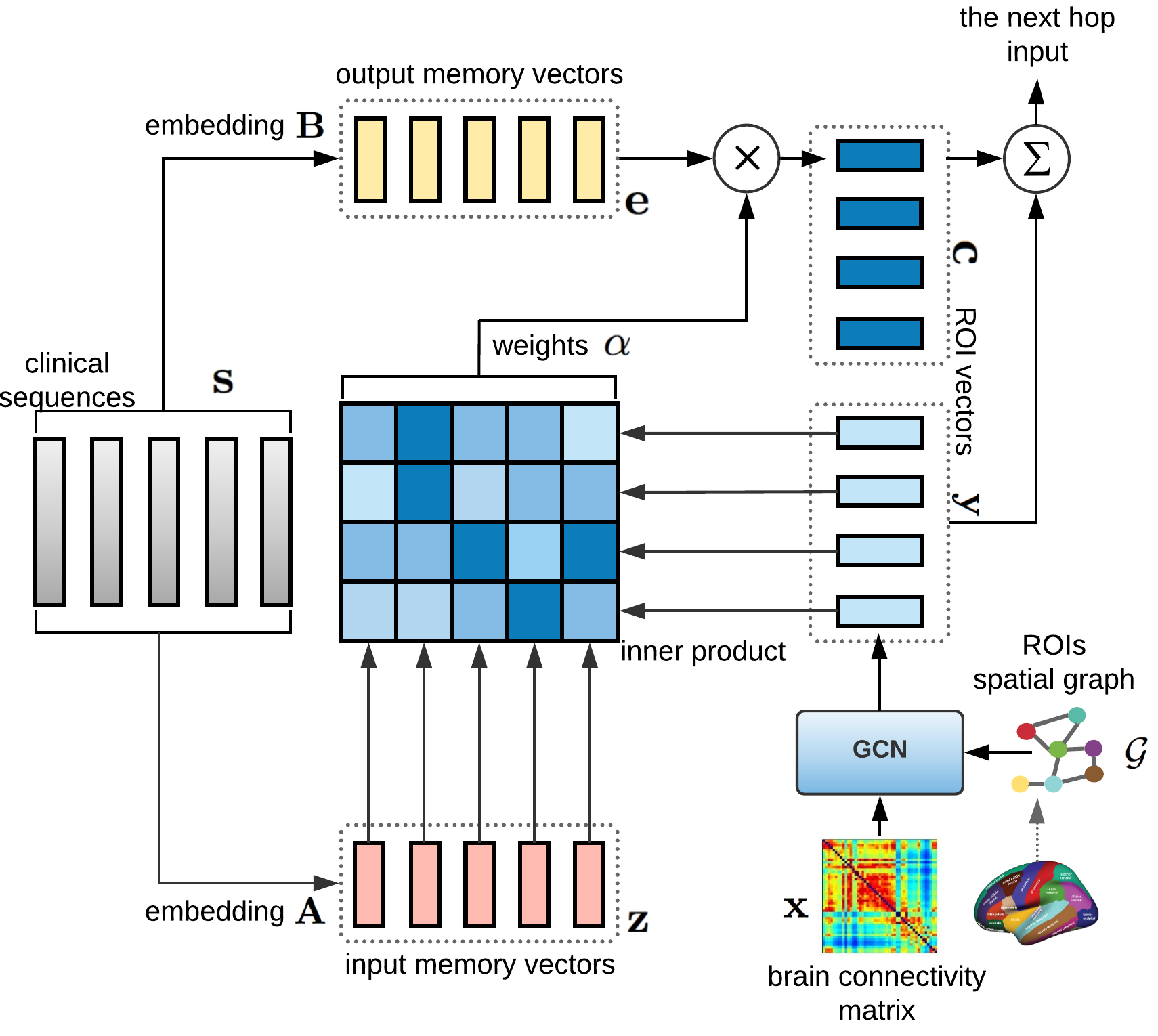}
    \caption{Illustration of memory augmented graph convolution in a single hop (the $1$-st hop). See Section~\ref{sec:memory} for the details.}
    \label{fig:memory}
\vspace{-0.3cm}
\end{figure}

\noindent{\bf Memory Representation Retrieving}

To retrieve memory vectors from the embedding space, we firstly need to decide which vector to choose. Not all records in a sequence contribute equally when it comes to the representation learning for brain graphs. Hence, attentive weights are adopted here to make a soft combination of all memory vectors. Mathematically, the weights are computed by a softmax function on the inner product of the input memory vectors $\mathbf{z}_{j}$ and the learned ROI vectors $\mathbf{y}_{i}$,
\begin{equation} \label{eq:attention}
\alpha_{ij}=softmax(\mathbf{y}_{i}\mathbf{z}_{j})=\frac{\exp(\mathbf{y}_{i}\mathbf{z}_{j})}{\sum_{j'=1}^{t}\exp(\mathbf{y}_{i}\mathbf{z}_{j'})}
\end{equation}
Once the informative memory vectors are indicated by weights $\alpha_{ij}$, the correspondence strength for attention are shown. As Fig.~\ref{fig:memory} illustrated, our attention is 2-dimensional that describes similarities between the representations generated from two modality sources. To make this feasible, we assume that both memory and ROI vectors are in the embedding space with same dimension. Next we represent the contextual information by the aggregation of weights and output memory vectors. Specifically, 
\begin{equation} \label{eq:context}
\mathbf{c}_{i}=\sum_{j=1}^{t}\alpha_{ij}\mathbf{e}_{i} 
\end{equation}
where $\mathbf{c}_{i}$ is a row vector of the context matrix $\mathbf{c}$, and is aware of a new representation for the ROI. 

To integrate the context vectors with feature maps of GCN, element-wise sum is employed as $\hat{\mathbf{y}}_{i}=\mathbf{y}_{i}+\mathbf{c}_{i}$. The intuition of using the sum operator derives from the neural networks~\cite{gehring2017convolutional}, in which the learned features in the next layer would benefit from both components in their networks. 

The entire operations in a single hop are shown in Fig.~\ref{fig:memory}, which is regarded as one layer (hop) of our model MemGCN. The output feature matrix of the single hop $\hat{\mathbf{y}}=(\hat{\mathbf{y}}_{1},\cdots,\hat{\mathbf{y}}_{n})$ is fed into the next hop, and again as an input of the next GCN.         

\subsection{Multi-hop Layer}
Basically, memory mechanism allow the network to read the input sequences multiple times to update the memory contents at each step and then make a final output. Compared to single step attention~\cite{bahdanau2014neural, rush2015neural}, contextual information from the memory is collected iteratively and cumulatively for feature maps learning. In particular, suppose there are $L$ layer memories for $L$ hop operations, the output feature map $\hat{\mathbf{y}}$ at the $l$-th hop can be rewritten as
\begin{equation}
\mathbf{y}^{l+1}=\mathbf{H}\mathbf{y}^{l}+\mathbf{c}^{l}, l=1,\cdots,L
\end{equation}
where $\mathbf{H}$ is a linear mapping and can be beneficial to the iteratively updating of $\mathbf{y}$. Similarly, the computational equations for weights and context vectors in Eq.(\ref{eq:attention}) and Eq.(\ref{eq:context}) are rewritten as
\begin{align} 
&\alpha_{ij}^{l}=\frac{\exp(\mathbf{y}_{i}^{l}\mathbf{z}_{j}^{l})}{\sum_{j'=1}^{t}\exp(\mathbf{y}_{i}^{l}\mathbf{z}_{j'}^{l})} \\
&\mathbf{c}_{i}^{l}=\sum_{j=1}^{t}\alpha_{ij}^{l}\mathbf{e}_{i}^{l} 
\end{align}
In addition, a layer-wise updating strategy~\cite{sukhbaatar2015end} for input and output memory vectors at multiple hops are used, which is keeping the same embeddings as $\mathbf{A}^{1}=\cdots=\mathbf{A}^{L}$ and $\mathbf{B}^{1}=\cdots=\mathbf{B}^{L}$. 

Notice that the contextual states $\mathbf{c}$ of the first hop are determined by the two given modalities and then accumulated into the generation of the contextual states in the following hops. Consequently, the final output feature maps $\mathbf{y}^{L}$ rely on the conditional contextual states $\mathbf{c}^{1},\cdots,\mathbf{c}^{L-1}$ as well as previous feature maps $\mathbf{y}^{1},\cdots,\mathbf{y}^{L-1}$, where $\mathbf{y}^{1}$ is directly generated from brain connectivity matrix $\mathbf{x}$ through one layer graph convolution. The underlying rationale of the multi-hop design is that it is easier for the model to learn what have already been taken into account in previous hops and capture a fine-grained attendance from memories in the current hop.     

\subsection{Matching Layer}
\label{sec:matching}
Metric learning for brain connectivity graphs with multiple layers normally involves several non-linearities so that the complex underlying data structure can be captured. To train such a neural network, a large amount of training data are necessary to prevent overfitting~\cite{bengio2009learning}. Although large-scale labeled dataset are often limited in clinical practice, metric learning between sample pairs allow us increase the training data significantly because of the possible combination of two samples~\cite{koch2015siamese}. In our case, take a brain image acquisition as a sample, the goal of metric learning is to learn discriminative properties to distinguish whether the sample pairs in the same diagnosis class or not. 

The basic hypothesis is that, if two samples share the same diagnosis result, the matching score between their high-level feature maps should be high. Here, two sorts of matching function are explored to calculate the similarities between pairs of acquisitions.  

\noindent{\bf Inner Product Matching}

Let $\mathbf{x}_{m}$ and $\mathbf{x}_{m'}$ denote any pair of initial brain connectivity matrices, $\mathbf{y}_{m,i}^{L}$ and $\mathbf{y}_{m',i}^{L}$ denote their associated feature vectors learned from the $L$-th hops by MemGCN, where $i$ is a vertex of ROI. The Euclidean distance computed in the matching layer is a vector with each dimension corresponding to each ROI, which is,
\begin{equation}
d_{i}(\mathbf{x}_{m},\mathbf{x}_{m'})=\|\mathbf{y}_{m,i}^{L}-\mathbf{y}_{m',i}^{L}\|_{2}, ~~i=1,\cdots,n
\end{equation}
Thus, $\mathbf{d}=(d_{1},\cdots,d_{n})$ is the output of the matching layer. Instead of computing the distance directly, the feature maps are normalized along with dimension of hidden features and then a inner product is used to get a similarity vector,   
\begin{equation} \label{eq:Inner}
    sim_{i}(\mathbf{x}_{m},\mathbf{x}_{m'}) = (\mathbf{y}_{m,i}^{L})^{\mathrm{T}}\mathbf{y}_{m',i}^{L},~~ i=1,\cdots,n.
\end{equation}
where $sim_{i}$ is the inner product similarity on $i$-th dimension, and it is equivalent to Euclidean distance if the vectors are normalized.   

\noindent{\bf Bilinear Matching}

Above matching function in Eq.(\ref{eq:Inner}) only considers the similarity of the corresponding ROI vectors of a given paired brain graphs. The similarities computed by different ROI are not modeled. To the aim, a simple bilinear matching function~\cite{yin2016neural} is used here. The matching score is defined as 
\begin{equation} \label{eq:Bilinear}
    sim_{i,j}(\mathbf{x}_{m},\mathbf{x}_{m'})=(\mathbf{y}_{m,i}^{L})^{\mathrm{T}}\mathbf{M} \mathbf{y}_{m',j}^{L},~~ i,j=1,\cdots,n.
\end{equation}
where $sim_{i,j}$ is the similarity between ROI $i$ and $j$ based on bilinear matching. $\mathbf{M}\in\mathbb{R}^{f^{L}_{out}\times f^{L}_{out}}$ is a matrix parameterizing the matching between the paired feature maps.
With the matching procedure in Eq.(\ref{eq:Bilinear}), the output of the matching layer is a matrix, with each element suggesting the strength of ROI connections. It is worth to note that if the parameter matrix $\mathbf{M}$ is an identity matrix, the bilinear matching reduces to the inner product matching.   

\subsection{Model Training}
As in MemGCN, our output layer models the probability of each sample pair is matching or non-matching. The similarity representation from matching layer is passed to a fully connected layer and a softmax layer for the eventual classification. For each pair, set the output of fully connected layer is a feature vector $\mathbf{r}$. We calculate the probability distribution over the binary classes by
\begin{equation} 
    \mathbf{p}=softmax(\mathbf{w}_{c}^{\mathrm{T}} \mathbf{r})
\end{equation}
where $\mathbf{w}_{c}\in\mathbb{R}^{2}$ is a trainable parameter. 

We train our model using a regularized cross-entropy loss function. Let $\mathcal{X}=\{(\mathbf{x}_{m}, \mathbf{x}_{m'})\}^{N}$ be the training set of $N$ acquisition pairs. $N$ is the number of total pairwise combination of brain graphs. The number of acquisitions $M$ is much smaller than $N$. The loss function we minimize is 
\begin{align} \label{eq:loss}
 \mathcal{L}= &\sum_{m,m'}^{N}\tilde{\mathbf{y}}_{m,m'}\log\mathbf{p}_{m,m'}
+(1-\tilde{\mathbf{y}}_{m,m'})\log(1-\mathbf{p}_{m,m'}) \nonumber \\
 &+ \gamma\|\mathbf{\Theta}\|_{2}
\end{align}
where $\tilde{\mathbf{y}}_{m,m'}$ denotes the label for sample pair $(\mathbf{x}_{m}, \mathbf{x}_{m'})$, $\mathbf{\Theta}$ is the collection of trainable parameters. The MemGCN is trained on machines with NVIDIA TESLA V100 GPUs by using Adam optimizer~\cite{kingma2014adam} with
mini-batch.

%% file: 3-experiment.tex
\section{Experiments}

\begin{table*}[t]
\centering
\tabcolsep 0.12in
\renewcommand\arraystretch{1.25}
\caption{Results for classifying matching vs. non-matching brain graphs on the test sets of tensor-FACT, ODF-RK2, and Hough in terms of Accuracy and AUC metrics. Performances without and with extra modalities are shown. ``Fusion'' modality means clinical records of both motor and non-motor features. (Hop number $L=3$ for MemGCNs).}
\label{tab:mainresults}
\begin{tabular}{l|l|cc|cc|cc}
\hline
\multirow{2}{*}{Extra Modalities}    & \multicolumn{1}{l|}{\multirow{2}{*}{Methods}} & \multicolumn{2}{c|}{tensor-FACT} & \multicolumn{2}{c|}{ODF-RK2} & \multicolumn{2}{c}{Hough}\\
                           & \multicolumn{1}{c|}{}                         & Accuracy          & AUC         & Accuracy        & AUC         & Accuracy        & AUC\\
\hline
\hline
\multirow{6}{*}{None}          & Raw Edges                                    &65.94 (3.78)                    &58.47 (4.05)              &67.56 (4.12)           &60.93 (5.60)              &67.90 (4.09)           &64.49 (3.56)    \\
                           & PCA                                          &69.19 (3.13)                   &64.10 (2.10)             &68.38 (2.50)                 &60.93 (2.63)              &66.28 (4.60)           &63.46 (3.52)        \\
                           & MLP                                 &84.22 (2.76)                &82.36 (2.87)        &82.31 (2.68)                       &82.53 (4.74)              &84.27 (2.63)           &81.77 (3.74)      \\
& GCN-$inner$                                        &93.69 (2.15)   &92.67 (4.94)               &93.23 (2.63)                  &93.04 (5.26) &92.80 (2.51)           &93.90 (5.48)      \\
&GCN-$bilinear$  &93.89 (1.76)   &94.77 (6.08)  &94.00 (2.65) &94.32 (5.72)   &93.34 (2.26)   &93.35 (5.14)      \\
\hline
\hline
\multirow{4}{*}{Fusion}       & AttGCN                                       &93.62 (2.99)                   &94.25 (5.88)             &94.76 (3.31)                 &94.33 (5.23)                        &94.01 (1.94)           &94.74 (5.35)   \\
& AttLstmGCN                                 &94.70 (2.35)                    &94.38 (5.41)             &94.89 (2.71)                 &94.87 (4.49)                        &94.64 (2.02)          &94.80 (5.51) \\
                           & MemGCN-$inner$                                 &95.43 (2.22)  &96.42 (6.36)   &95.54 (2.98)                 &96.59 (6.44)  &95.48 (2.34)           &96.49 (6.41)                 \\
                           & MemGCN-$bilinear$                                 &\textbf{95.47 (2.25)}&\textbf{96.48 (6.40)}   &\textbf{95.87 (2.56)}                 &\textbf{96.84 (6.36)}    &\textbf{95.64 (2.00)}  &\textbf{96.74 (6.51)}   \\
\hline
\end{tabular}
\vspace{-0.3cm}
\end{table*}

\subsection{Dataset}
The data we used to evaluate MemGCN are obtained from the Parkinson Progression Marker Initiative (PPMI)~\cite{marek2011parkinson} study. PPMI is an ongoing PD study that has meticulously collected various potential PD progression markers that have been conducted for more than six years. Neuroimages and EHRs are considered as two modalities in this work.

To obtain brain connectivity graphs, a series of preprocessing procedures are conducted. For the correction of head motion and eddy current distortions, FSL eddy-correct tool is used to align the raw data to the b0 image. Also, the gradient table is corrected accordingly. To remove the non-brain tissue from the diffusion MRI, the Brain Extraction Tool (BET) from FSL~\cite{smith2002fast} is used. To correct for echo-planar induced (EPI) susceptibility artifacts, which can cause distortions at tissue-fluid interfaces, skull-stripped b0 images are linearly aligned and then elastically registered to their respective preprocessed structural MRI using the Advanced Normalization Tools (ANTs\footnote{\url{http://stnava.github.io/ANTs/}}) with SyN nonlinear registration algorithm~\cite{avants2008symmetric}.  The resulting 3D deformation fields are then applied to the remaining diffusion-weighted volumes to generate full preprocessed diffusion MRI dataset for the brain network reconstruction. In the meantime, ROIs are parcellated from T1-weighted structural MRI using Freesufer\footnote{\url{https://surfer.nmr.mgh.harvard.edu}}. 

The connectivity graphs computed by three whole brain tractography methods~\cite{zhan2015comparison} for are applied, which is a coverage of the tensor-based deterministic approach (Fiber Assignment by Continuous Tracking~\cite{mori1999three}), the Orientation Distribution Function (ODF)-based deterministic approach (the 2nd-order Runge-Kutta, RK2~\cite{basser2000vivo}), as well as the probabilistic approach (Hough voting~\cite{aganj2011hough}). $84$ ROIs are finally obtained. 
We define each the coordinates for ROIs using the mean coordinate for all voxels in the corresponding regions (see Spatial Graph Construction in Section~\ref{sec:gcn} for the details). After preprocessing, we collect a dataset of $754$ Diffusion Tensor Imaging (DTI) acquisitions, where $596$ of them are brain graphs of Parkinson's Disease (PD) patients and the rest $158$ are from Healthy Control (HC) subjects. The spatial graph we constructed has $84$ vertices and $527$ edges, with each vertex is corresponding to an ROI.

Additionally, sequential EHR records are aligned with corresponding brain connectivity graphs. For each acquisition $\mathbf{x}$, a sequence of its associated input features $(\mathbf{s}_{1},\cdots,\mathbf{s}_{t})$ can be used for the external memories. Note that the sequences are chunked at the time points of neruoimaging acquisition, and we only use the subsequences before the time point to make a reasonable experimental design. Usually, the number of timestamps in sequences are different because subjects provide their medical records with distinct frequencies, we set the length of sequence as $t=12$ according to the statistics of the PPMI study. Padding is utilized for those sequences with fewer timestamps. The specific clinical assessments we study here are motor (MDS-UPDRS Part II-III~\cite{goetz2008movement}) and non-motor (MDS-UPDRS Part I~\cite{goetz2008movement} and MoCA~\cite{nasreddine2005montreal}) symptoms which are crucial for evaluating a disease course of PD. There are $79$ discrete clinical features and $331$ dimensions after binarization, then we have the original dimensions of clinical feature which is $D=331$. 


\subsection{Experimental Setup}
\noindent{\bf Implementation Details}

To learn similarities between brain connectivity matrices, acquisitions in the same group (PD or HC) are labeled as matching pairs while those from different groups are labeled as non-matching pairs. Hence, we have $283,881$ pairs in total, with $189,713$ matching pairs and $94,168$ non-matching pairs. 

We selected hyperparameter values through random search~\cite{bergstra2012random}. Batch size is $32$. Initial learning rate is $5e$-$3$, and early stop is used once the model stops improving. The L2-regularization weight is $1e$-$2$. 
For each graph convolution operation, the order of Chebyshev polynomials and the feature map dimension are respectively set as $r=30$ and $f_{out}=32$. For the memory network, memory size and dimension of embedding are respectively set as $t=12$ and $d=32$. The code is available at~\url{https://github.com/sheryl-ai/MemGCN}.

\begin{table*}[t]
\centering
\tabcolsep 0.158in
\renewcommand\arraystretch{1.25}
\caption{Comparisons for MemGCN by various setting of the number of hops and the matching methods.}
\label{tab:multihops}
\begin{tabular}{c|l|cc|cc|cc}
\hline
\multirow{2}{*}{\# of hops} & \multicolumn{1}{l|}{\multirow{2}{*}{Matching}} & \multicolumn{2}{c|}{tensor-FACT} & \multicolumn{2}{c|}{ODF-RK2} & \multicolumn{2}{c}{Hough}\\
& & Accuracy          & AUC         & Accuracy        & AUC         & Accuracy        & AUC\\
\hline
\hline
1 &inner product  &94.20 (2.42)  &94.07 (5.19) &94.03 (2.32) &94.39 (5.15) &94.61 (2.05) &95.72 (5.50) \\
2  &inner product  &95.36 (2.60)  &96.35 (6.36) &95.40 (2.27) &96.39 (6.35) &95.21 (2.92)  &96.10 (6.30) \\
3  &inner product    &\textbf{95.43 (2.22)}  &\textbf{96.42 (6.36)}   &\textbf{95.54 (2.98)}      &\textbf{96.59 (6.44)}     &\textbf{95.48 (2.34)}  &\textbf{96.49 (6.41)} \\  
\hline
1 &bilinear  &94.68 (2.04) &95.32 (5.98)  &93.88 (2.22)               &94.17 (5.49)  &94.37 (1.89)   &95.17 (4.95)\\
2 &bilinear  &95.19 (2.14) &96.06 (6.18)       &94.61 (2.91)               &95.27 (5.89)   &95.23 (2.50)   &96.17 (5.26)\\  
3 &bilinear  &\textbf{95.47 (2.25)} &\textbf{96.48 (6.40)}                 &\textbf{95.87 (2.56)}   &\textbf{96.84 (6.36)}   &\textbf{95.64 (2.00)}     &\textbf{96.74 (6.51)} \\ 
\hline
\end{tabular}
\vspace{-0.3cm}
\end{table*}

\noindent{\bf Baselines}

To test the performance of MemGCN, we report the empirical results of comparisons with a set of baselines. Here are the methods that classify brain graphs without any other modalities.    

\noindent{\bf $\bullet$ \emph{Raw Edges.}} It is one simple approach that is to directly use the numerical values from the connectivity matrix to represent the brain network. The feature space is a $84\times 84$ vector.  

\noindent{\bf $\bullet$ \emph{PCA.}} Principal Component Analysis (PCA) is used to reduce the data dimensionality. After forming a sample-by-feature input matrix, PCA is performed by keeping the first 100 principal components, which is an optimal setting in practice. 

\noindent{\bf $\bullet$ \emph{MLP.}} Multilayer perceptron (MLP) is employed. The hidden dimension of the first layer is set as $f_{out}=1024$. We use a $3$-layers fully connected network, with its output layer reduces the dimension down from $64$ to $2$ by a softmax. It previously used on brain graphs in~\cite{munsell2015evaluation}.

\noindent{\bf $\bullet$ \emph{GCN-inner.}} Metric Learning for brain networks using GCN is first introduced in~\cite{ktena2017distance}, where a global loss function is used to supervise pairwise similarities. The cross-entropy loss is adopted in our experiments to be consistent with other models.

\noindent{\bf $\bullet$ \emph{GCN-bilinear.}} The bilinear matching layer proposed in Section~\ref{sec:matching} is added on the basis of GCN to conduct metric learning. It is a version of MemGCN-bilinear without memory.   

Furthermore, neural networks without memory mechanism that can embed the clinical data are built as baselines.   

\noindent{\bf $\bullet$ \emph{AttGCN.}} Instead of using input and output memories on sequences, only one embedding matrix for the computation of attentive weights is incorporated with GCN via the sum operation.    

\noindent{\bf $\bullet$ \emph{AttLstmGCN.}} A standard bi-directional LSTM with attention~\cite{bahdanau2014neural} is established for sequential EHR data and then its context states are combined with GCN feature maps.    

Finally, two variants of our model are given.

\noindent{\bf $\bullet$ \emph{MemGCN-inner.}} The proposed MemGCN with the inner product matching layer. 

\noindent{\bf $\bullet$ \emph{MemGCN-bilinear.}} The proposed MemGCN with the bilinear matching layer. 
     
For a fair comparison, the reported models are built under the same pairwise matching architecture for metric learning. Inner product matching are employed in the baseline model if it is not stated as a bilinear version. $5$-fold cross validation on $754$ DTI acquisitions (1-fold for generating held-out testing pairs) are conducted in all of our experiments.  

\begin{table*}[t]
\centering
\tabcolsep 0.112in
\renewcommand\arraystretch{1.25}
\begin{threeparttable}
\caption{The interpretability of the output representation of MemGCN's inner product matching layer. Top-$5$ identical ROIs in PD group and discriminative ROIs between PD and HC groups are listed. Similarity scores are given.}
\label{tab:hopmatching}
\label{tab:innerinterpretability}
\begin{tabular}{|l|l|c|l|c|l|c|}
\hline
\multicolumn{1}{|l|}{\multirow{2}{*}{}}                                                          & \multicolumn{2}{c|}{Motor}                             & \multicolumn{2}{c|}{Non-motor}                         & \multicolumn{2}{c|}{Fusion}                            \\ \cline{2-7} 
\multicolumn{1}{|l|}{}                                                                           & \multicolumn{1}{c|}{ROI Name} & \multicolumn{1}{c|}{Score} & \multicolumn{1}{c|}{ROI Name} & \multicolumn{1}{c|}{Score} & \multicolumn{1}{c|}{ROI Name} & \multicolumn{1}{c|}{Score} \\ 
\hline
\multirow{5}{*}{\begin{tabular}[c]{@{}l@{}}Identical ROIs\\ (PD Group)\end{tabular}}             &Right Thalamus Proper                           &0.9258                            &Rh Paracentral             &0.8563          &Rh Pars Opercularis                             &0.9344                            \\
                                                                                                 &Lh Insula                            &0.9253                            &Rh Lingual                           &0.8180                            &Rh Lateral Occipital                          &0.8372                            \\
                                                                                                 &Right Pallidum                           &0.9226                            &Right Pallidum                           &0.8091                            &Left Accumbens Area                           &0.7887                            \\
                                                                                                 &Lh Rostral Middle Frontal                            &0.9210                            &Lh Parsorbitalis                           &0.6554                            &Rh Parahippocampal                           &0.7827                            \\
                                                                                                 &Parahippocampal                          &0.9206                            &Left Thalamus Proper                           &0.6387                            &Rh Frontalpole                            &0.7742                            \\
                                                                                   
                                                                                   \hline
\multirow{5}{*}{\begin{tabular}[c]{@{}l@{}}Discriminative ROIs\\ (PD vs. HC Group)\end{tabular}} &Right Putamen                           &-0.9134                            &Left Putamen                           &-0.7423                            &Right Thalamus Proper                          &-0.8960                             \\
                                                                   &Right Accumbens Area                           &-0.9075                            &Lh Frontal Pole                              &-0.5754                         &Left Caudate                          &-0.8439                           \\
                                                                     &Left Hippocampus                           &-0.9059                            &Lh Supramarginal                            & -0.5731                            &Lh Paracentral                           &-0.8227                            \\
                                                                     &Right VentralDC                            &-0.9058                            &Lh Inferior Parietal                          &-0.5693                            &Lh Middle Temporal                           &-0.7865                            \\
                                                                     &Left Caudate                           &-0.9014                            &Lh Paracentral                           &-0.4851                            &Lh Cuneus                            &-0.7528       \\
                                                                                   \hline                  
\end{tabular}
  \begin{tablenotes}
    \small
    \item $^*$ Lh and Rh are the abbreviations of Left Hemisphere and Right Hemisphere respectively.
  \end{tablenotes}
\end{threeparttable}
\end{table*}

\subsection{Results}
\noindent{\bf Matching vs. Non-matching Classification}

Table~\ref{tab:mainresults} reports the performance for the binary classification task. The metrics for evaluation are Accuracy and Area Under the Curve (AUC).  

From the results we can observe that, the Raw Edges, and simple feature extraction approach such as PCA cannot predict a reliable distance for sample pairs and correspondingly achieve promising results on the matching classification task. More layers with extra non-linearities have a good influence on the fully connected networks to capture the complicated patterns from acquisitions. All the GCN based methods can largely improve both of Accuracy and AUC performance in three DTI sets generated by Tensor-FACT, ODF-RK2, and Hough tractography algorithms, which demonstrates the effectiveness of graph convolution on the brain connectivity graphs. Overall, the bilinear matching strategy is outperform the inner product matching strategy slightly on both GCN and MemGCN. The best AUC performance is $96.48$, $96.84$, and $96.74$, which are accomplished by MemGCN-bilinear with fusion clinical sequences as the external modality. With attention mechanism, AttGCN and AttLstmGCN also perform well in the given circumstances. However, they cannot boost the results significantly compared to the vanilla GCN. The reason that MemGCN behaves better than them is probably separate memories for reading and retrieving are employed in a multi-hop network.   

Table~\ref{tab:hopmatching} shows the concrete effects of increasing the number of hops on inner product and bilinear matchings. The number of hops is tuned from $1$ to $3$. The results on Accuracy and AUC metrics illustrate that our multi-hop framework indeed improves performance constantly. 

\noindent{\bf Identical ROIs vs. Discriminative ROIs}

The interpretability of MemGCN is investigated. As the representation learned in the inner product matching layer can be explained as pairwise similarities at $84$ ROI dimensions, it describes the significance of each ROI in metric learning. Therefore, we compute the average similarities for all the PD-PD pairs and the average similarities for all the PD-HC pairs. ROIs with the highest scores in the PD group could be considered as the identical ROIs for PD, while those with the lowest scores in the PD versus HC group are regarded as the discriminative ROIs. 

The interpretable results depends on memory augmentation of motor, non-motor, and the fusion data are presented in Table~\ref{tab:innerinterpretability}. While the whole functions of the human brain regions are still unclear, it is quite intriguing that MemGCN can locate some of the modality-related ROIs, which might be critical for PD study. For instance, The most identical ROI for PD with motor features as augmentation is the Thalamus with one of its major role as motor control. Also, lingual gyrus discovered by non-motor features is linked to processing vision, especially related to letters. On the other hand, MemGCN can help us to find which ROI is sufficiently discriminative to distinguish PD patients with healthy controls. Several important ROIs belongs to the current research of clinicians and domain experts are detected, i.e., Caudate and Putamen areas.   

To show the representation generated from the bilinear matching layer, we draw the edges between ROIs with high similarities in Fig.3. Similar to Table~\ref{tab:innerinterpretability}, the most identical edges for PD group and the most discriminative edges between PD and HC groups are depicted. The interesting patterns we found might be deserved to the further exploration in clinical scenarios.

\begin{figure}[t]
\subfigure[]{
\begin{minipage}[b]{0.5\linewidth}
\centering 
\includegraphics[width=\textwidth]{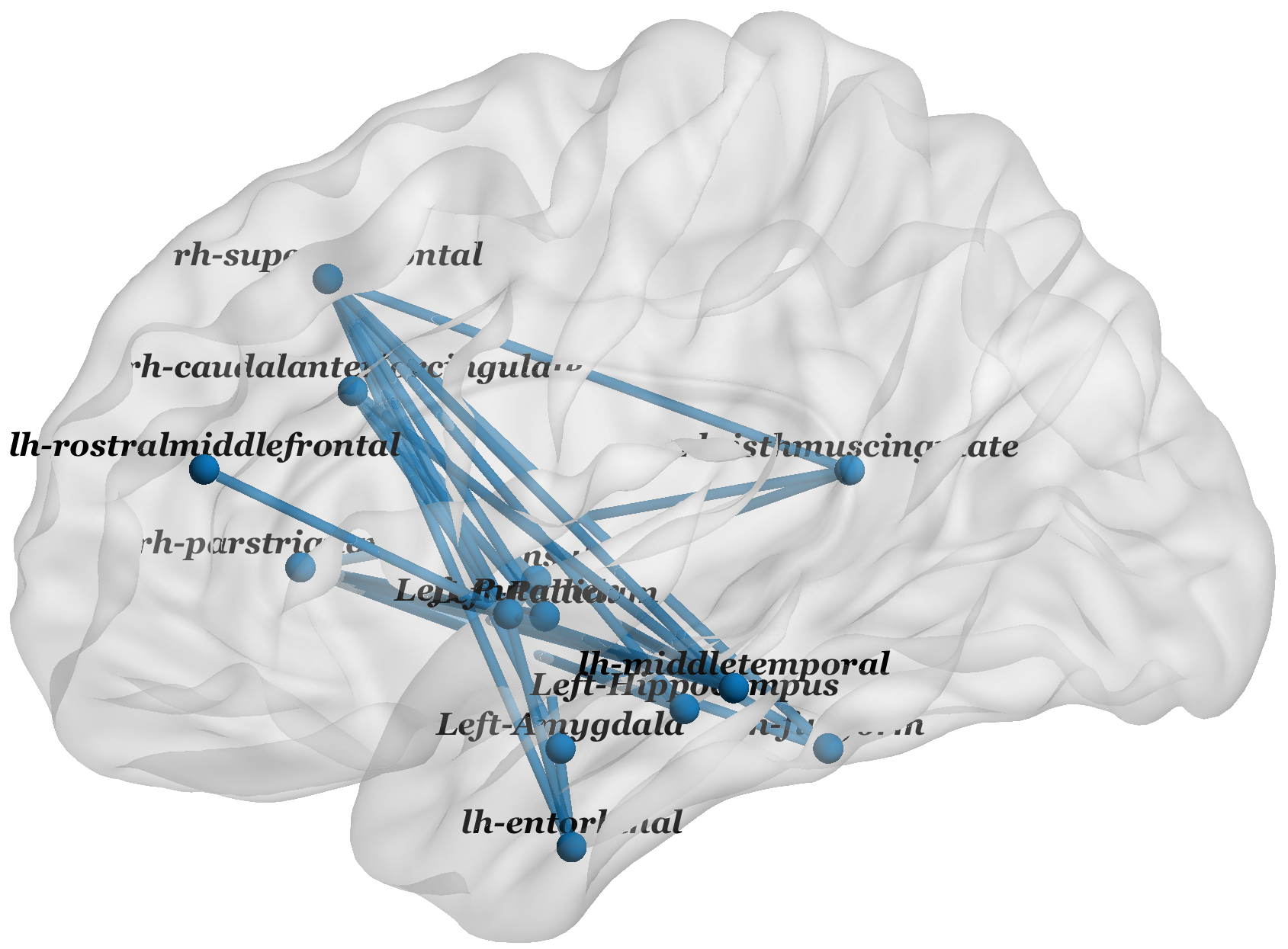}
\end{minipage}}%
\subfigure[]{
\begin{minipage}[b]{0.5\linewidth}
\centering
\includegraphics[width=\textwidth]{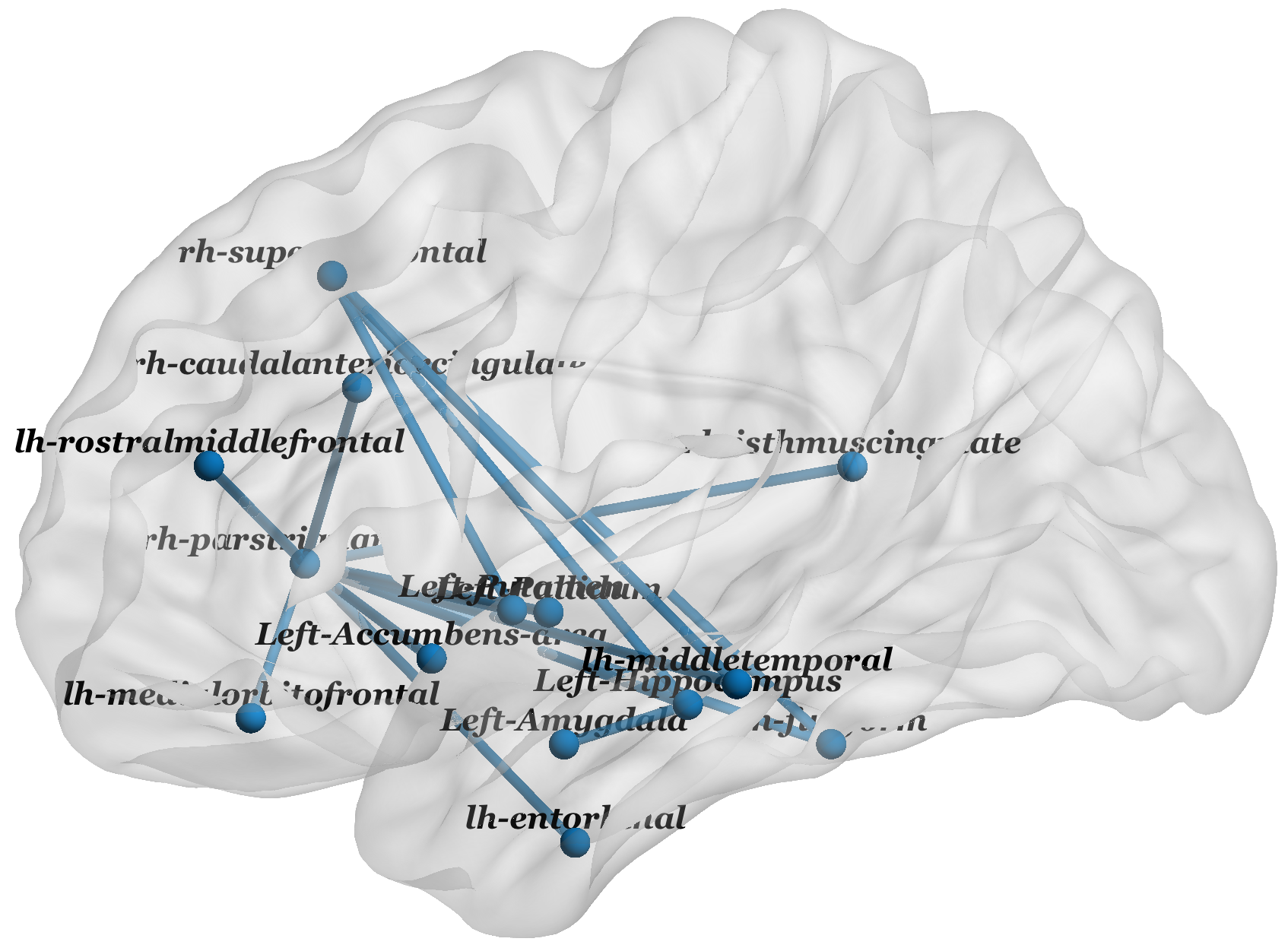}
\end{minipage}}
\vspace{-0.5cm}
\caption{The connectivity patterns learned by the bilinear matching layer. (a) The top identical edges for PD group; (b) The top discriminative edges between PD and HC groups.}
\vspace{-0.3cm}
\end{figure}

\begin{figure*}[t]
\subfigure[]{
\begin{minipage}[b]{0.5\linewidth}
\centering 
\label{fig:pd}
\includegraphics[width=3in]{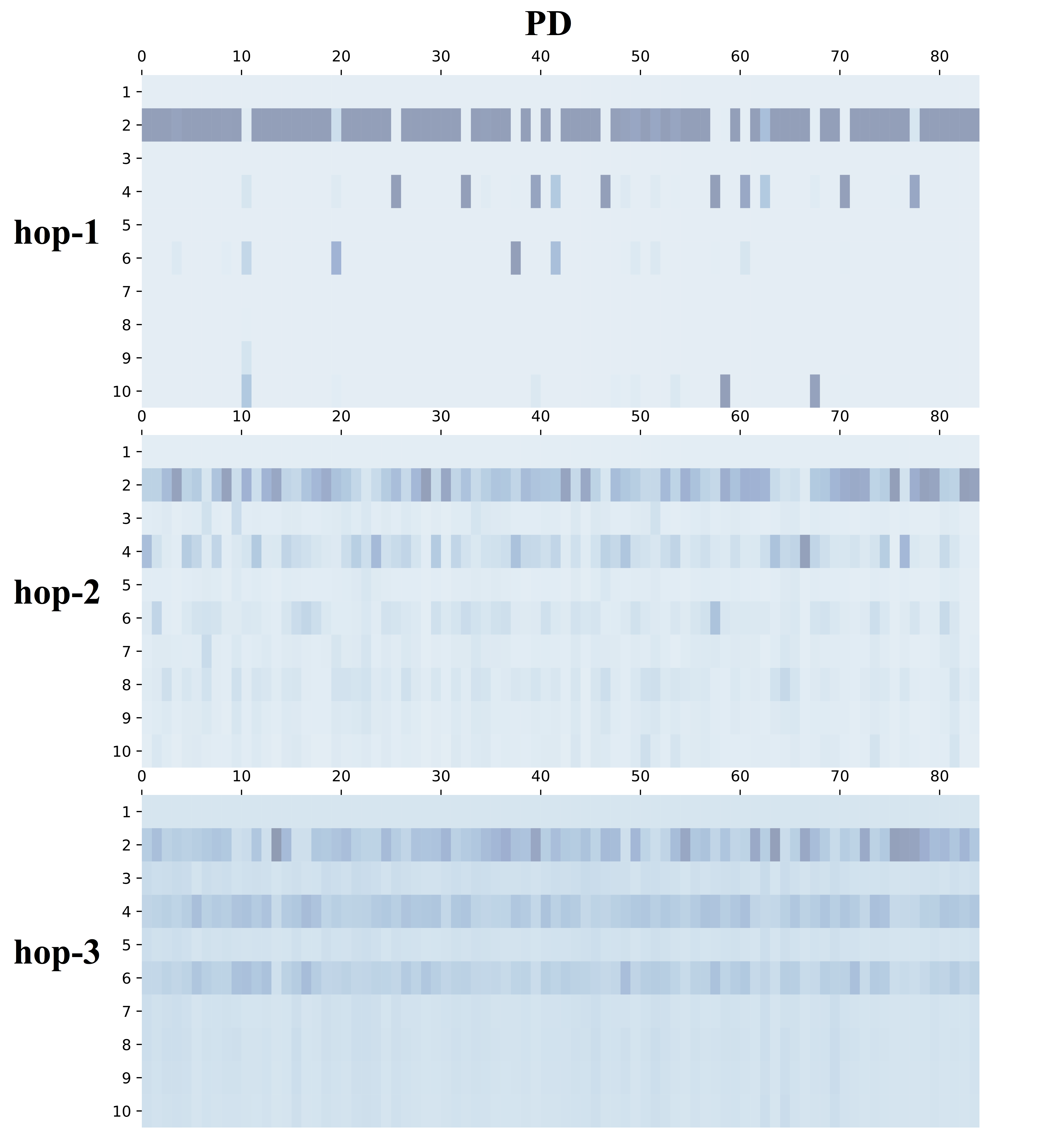}
\end{minipage}}%
\subfigure[]{
\begin{minipage}[b]{0.5\linewidth}
\centering
\label{fig:hc}
\includegraphics[width=3in]{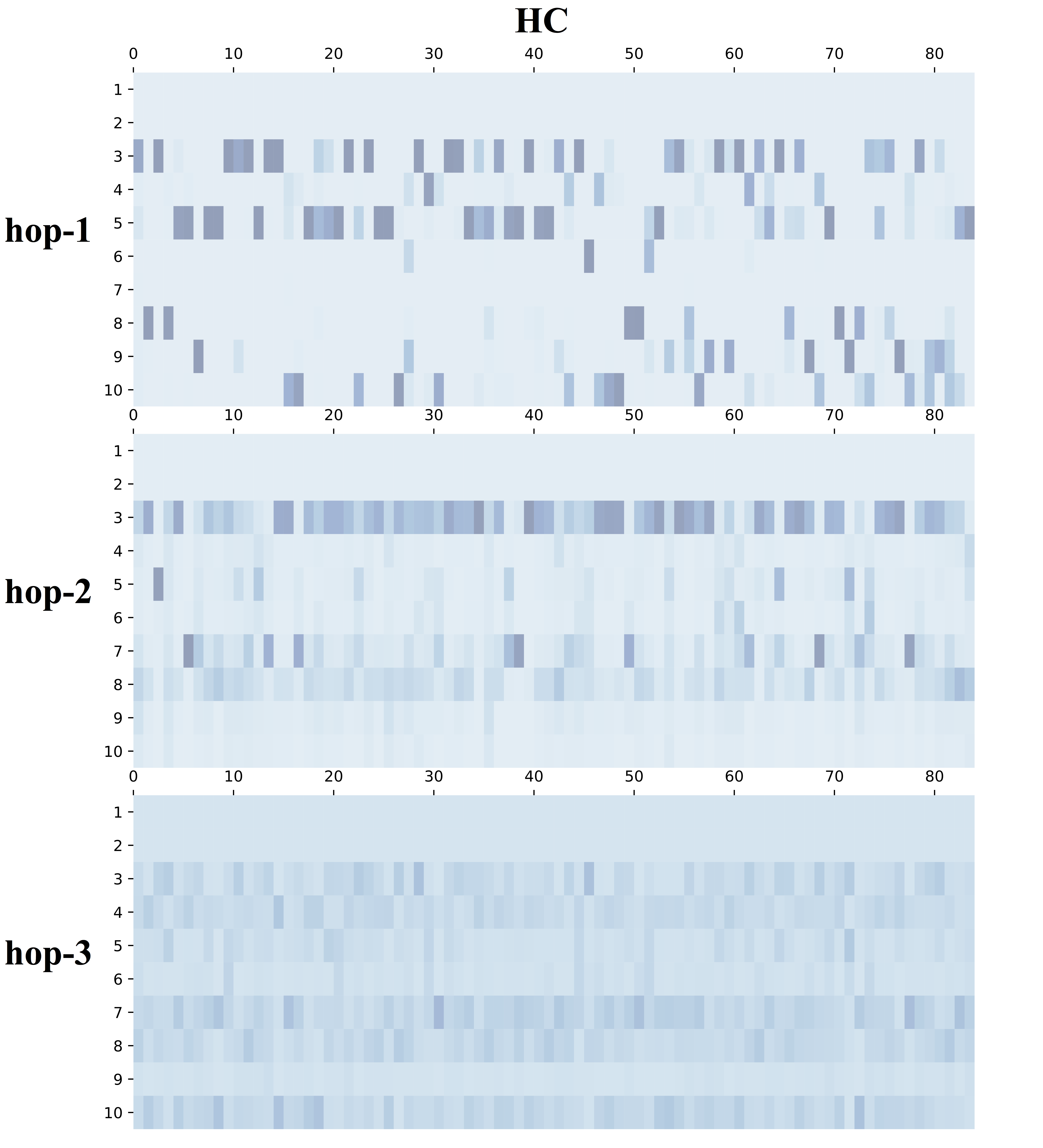}
\end{minipage}}
\vspace{-0.5cm}

\caption{Visualizations of attention interaction matrices of MemGCN for one PD and one HC case during $3$ memory hops. The rows and columns of the matrices respectively denote memory positions and ROI vertices. The darker color in the colormaps means a larger value which is close to 1, and the lighter color means a smaller value which is close to 0.}
\vspace{-0.3cm}
\end{figure*}

\noindent{\bf Longitudinal Alignment: Case Study}

From Fig.~\ref{fig:pd} and~\ref{fig:hc} we observe that though the structures of three hops of memory layer are same, the values of the attention weights they learned are quite different in typical cases. The matrices we draw in terms of colormaps in Fig.4 indicate the attentive weights $\alpha$ for one PD case and one healthy control case. Here we abandon the first $2$ padding dimensions of the shown cases and give $10$ memory positions (rows in the matrices). The attendance of all the $84$ ROI vertices are depicted (columns in the matrices). A darker color indicates where MemGCN is attending during the multi-hop updating for representations. Basically, given a specific case, which time point has more influences on his/her PD progression and which ROI is more important according to the clinical evidences can be analyzed through this longitudinal alignment between DTIs and EHRs.     

In general, the first hop attention appears to be primarily concerned with identifying the salient interaction between time-aware sequences and ROIs' feature maps. In this hop, the majority values are close to zero and only a few values are close to one, such that a sketch of key ROIs and timestamps are signified. The second and the third hops are then responsible for the fine-grained interactions that are relevant to optimizing the representation for the distance learning task.  

Another important observation is that the PD case has different interaction patterns compared to the healthy control. At each hop, PD has a relatively narrow attention and fewer responses across memory positions. Consider the PD case shown in Fig.~\ref{fig:pd}, longitudinal alignments occur at timestamps $2$, $4$, and $6$ after 3-hop updating, meanwhile a series of ROIs might function on the disease progression. By the Desikan-Killiany Atlas, the darker ROI dimensions from $76$ to $79$ are Rh Insula, Right Thalamus Proper, Right Caudate, and Right Putamen, respectively, which matches our expectation for the PD case.

%% file: 4-relatedwork.tex
\section{Related Work}

We briefly review the existing research that is closely related to the framework proposed in this paper.

{\bf EHR Mining}. In recent years many algorithms have been proposed to mine insights from patient EHRs. Initially those methods were static in the sense that they first construct patient vectors by aggregating their EHR with in a certain observational time window and then build learning approaches (e.g., predictive models and clustering methods) on top of those vectors \cite{caruana2015intelligible,ghassemi2014unfolding}. Most of these methods are shallow except the DeepPatient work which applied AutoEncoder to further compress the patient vectors and obtain better representations \cite{miotto2016deep}. Recently researchers have also been exploring CNN and RNN type of approaches to incorporate the temporal information in patient EHRs into the modeling process \cite{che2017rnn,cheng2016risk,ma2018health}. However, these methods compressed the patient EHRs to a vector before it was fed to the final model, which is not as flexible as the memory network we adopted.

{\bf GCN for Neuroimage Analysis}. Many data mining approaches have been developed to perform neuroimage analysis in recent years \cite{megalooikonomou2000data}, among which deep learning models are very popular because of their huge success in various computer vision problems \cite{litjens2017survey}. Recently, Ktena {\em et al.} \cite{ktena2017distance} propose to learn a metric from patients' neuroimages on top of the features constructed using GCN (where the graph is basically the patients' brain network constructed on the ROIs), which can discriminate the cases versus controls with autism. Zhang {\em et al.} \cite{zhang2018multi} extend such approach to handle the multiple modalities of the brain networks (e.g., constructed from different tractography algorithms on DTI images). However, none of them incorporated any clinical records from the patients. Our work is the first step towards filling the gap.

%% file: 5-conclusion.tex
\section{Conclusion} 

We propose a novel framework, memory-based graph convolution network (MemGCN), to perform integrative analysis of patient clinical records and neuroimages. On the one hand, our experiments on classification of Parkinson's Disease case patients with healthy controls demonstrate the superiority of MemGCN over conventional approaches. On the other hand, the interpretable high-level representations extracted from the inner product or bilinear matching layers are capable of indicating group patterns of brain connectivity via ROI nodes or their edges for PD subjects and healthy controls. 

Here we explored the operators of the graph convolution via ChebNet and the embedding via memory mechanism as feature extractors for neuroimages and patient health records respectively. The pairwise distance under the metric learning setting in our framework makes a progress in modeling a small cohort data such as PPMI. An important future direction is to design deep architectures that can lower the amount of training data meanwhile learn meaningful representations. We are especially interested in continuing to develop more general end-to-end trainable models in the space of boosting system performance on small data.

%% file: 6-acknowledgment.tex
\section*{Acknowledgement}

The authors would like to thank Dr. Liang Zhan's help on processing the neuroimages. The research is supported by NSF IIS-1716432, NSF IIS-1750326, and Michael J. Fox Foundation grant number 14858. Data used in the preparation of this article were obtained from the Parkinson's Progression Markers Initiative (PPMI) database (\url{http://www.ppmi-info.org/data}). For up-to-date information on the study, visit \url{http://www.ppmi-info.org}. PPMI -- a public-private partnership -- is funded by the Michael J. Fox Foundation for Parkinson's Research and funding partners, including Abbvie, Avid, Biogen, Bristol-Mayers Squibb, Covance, GE, Genentech, GlaxoSmithKline, Lilly, Lundbeck, Merk, Meso Scale Discovery, Pfizer, Piramal, Roche, Sanofi, Servier, TEVA, UCB and Golub Capital. The authors would like to thank the support from Amazon Web Service Machine Learning for Research Award (AWS MLRA).